\documentclass{article}

\usepackage{arxiv}

\usepackage[utf8]{inputenc} 
\usepackage[T1]{fontenc}    
\usepackage[colorlinks=true, linkcolor=black, citecolor=black, urlcolor=black]{hyperref}       
\usepackage{url}            
\usepackage{booktabs}       
\usepackage{amsfonts}       
\usepackage{nicefrac}       
\usepackage{microtype}      
\usepackage{lipsum}		
\usepackage{graphicx}
\usepackage[numbers]{natbib}
\usepackage{doi}
\usepackage{amsmath}
\usepackage{amssymb}

\usepackage{enumitem}

\setlist[itemize]{leftmargin=20pt}

\title{BlazeAIoT: A Modular Multi-Layer Platform for Real-Time Distributed Robotics Across Edge, Fog, and Cloud Infrastructures}

\author{ \href{https://orcid.org/0009-0005-8194-2036}{\includegraphics[scale=0.06]{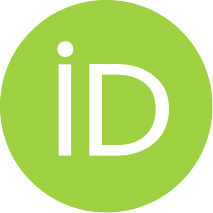}\hspace{1mm}Cédric Melançon} \\
	Department of Electrical Engineering\\
	École de technologie supérieure\\
	Montréal, Canada \\
	\texttt{cedric.melancon.1@ens.etsmtl.ca} \\
	\And
	\href{https://orcid.org/0000-0002-4091-3790}{\includegraphics[scale=0.06]{orcid.pdf}\hspace{1mm}Julien Gascon-Samson} \\
	Department of Software \& IT Engineering\\
	École de technologie supérieure\\
	Montréal, Canada \\
	\texttt{julien.gascon-samson@etsmtl.ca} \\
	\And
    \href{https://orcid.org/0000-0003-2547-2509}{\includegraphics[scale=0.06]{orcid.pdf}\hspace{1mm}Maarouf Saad} \\
	Department of Electrical Engineering\\
	École de technologie supérieure\\
	Montréal, Canada \\
	\texttt{maarouf.saad@etsmtl.ca} \\
    \And
    \href{https://orcid.org/0000-0003-4597-1700}{\includegraphics[scale=0.06]{orcid.pdf}\hspace{1mm}Kuljeet Kaur} \\
	Department of Electrical Engineering\\
	École de technologie supérieure\\
	Montréal, Canada \\
	\texttt{kuljeet.kaur@etsmtl.ca} \\
    \And
    Simon Savard \\
	Department of Electrical Engineering\\
	École de technologie supérieure\\
	Montréal, Canada \\
	\texttt{simon.savard.2@ens.etsmtl.ca} \\
}

\renewcommand{\shorttitle}{BlazeAIoT Real-Time Distributed Robotics Platform}

\hypersetup{
pdftitle={BlazeAIoT: A Modular Multi-Layer Platform for Real-Time Distributed Robotics Across Edge, Fog, and Cloud Infrastructures},
pdfsubject={cs.RO, cs.DC},
pdfauthor={Cédric Melançon, Julien Gascon-Samson, Maarouf Saad, Kuljeet Kaur, and Simon Savard},
pdfkeywords={Adaptive communication protocols, Cloud robotics, Dynamic data distribution, Distributed robotics, Edge robotics, Fog robotics, IoT platform.},
}

\begin{document}
\maketitle

\begin{abstract}
	The increasing complexity of distributed robotics has driven the need for platforms that seamlessly integrate edge, fog, and cloud computing layers while meeting strict real-time constraints. This paper introduces BlazeAIoT, a modular multi-layer platform designed to unify distributed robotics across heterogeneous infrastructures. BlazeAIoT provides dynamic data transfer, configurable services, and integrated monitoring, while ensuring resilience, security, and programming language flexibility. The architecture leverages Kubernetes-based clusters, broker interoperability (DDS, Kafka, Redis, and ROS2), and adaptive data distribution mechanisms to optimize communication and computation across diverse environments. The proposed solution includes a multi-layer configuration service, dynamic and adaptive data bridging, and hierarchical rate limiting to handle large messages. The platform is validated through robotics scenarios involving navigation and artificial intelligence-driven large-scale message processing, demonstrating robust performance under real-time constraints. Results highlight BlazeAIoT’s ability to dynamically allocate services across incomplete topologies, maintain system health, and minimize latency, making it a cost-aware, scalable solution for robotics and broader IoT applications, such as smart cities and smart factories.
\end{abstract}

\keywords{Adaptive communication protocols \and Cloud robotics \and Dynamic data distribution \and Distributed robotics \and Edge robotics \and Fog robotics \and IoT platform.}

\twocolumn
\section{Introduction}
The distributed robotics concept first emerged with the advent of cloud robotics \cite{Quintas2011CloudNetworks} and has since evolved into edge robotics \cite{Antevski2018EnhancingInformation} and fog robotics \cite{KrishnaChandGudi2018FogInteraction}. At its core, distributed computing in robotics addresses the limitations of onboard resources and the restricted energy autonomy imposed by battery-powered systems. Offloading intensive computations to external layers alleviates these constraints, but simultaneously introduces challenges related to network connectivity \cite{Sahni2017EdgeThings}, latency \cite{Bedard2023MessageSystems}, and the complexity of distributed topologies \cite{Lindsay2021TheFrontiers}

This work builds upon our previous work, a distributed data flow framework called BlazeFlow \cite{Melancon2023BlazeFlow:Applications}, which helps bridging data in a multi-layer distributed configuration. The proposed BlazeAIoT platform is introduced to encompass the three different concepts of Distributed Robotics, namely, edge, fog, and cloud robotics, in a single solution. To demonstrate its feasibility, a concrete robotics scenario is presented that evaluates the distribution of computational tasks across these layers and examines the resulting impact on the robot execution. Given that most robotics services operate under strict real-time constraints, particular attention is paid to assessing whether services can be reliably executed at each layer, or whether network latency and other factors render specific deployments impractical.

\subsection{Motivation}\label{section:motivation}
Although multiple distributed robotics platforms have been proposed in the literature, as described in Section \ref{section:related_work}, none support all three layers (edge, fog, and cloud) simultaneously. BlazeAIoT aims to introduce additional capabilities, as a platform, to address the challenges posed by such a distributed solution. The motivation for developing a novel Distributed Robotics platform is driven by multiple objectives:

\begin{itemize}
    \item \textbf{Cost-aware solution}: Minimize platform cost by leveraging open-source components and maintaining vendor-agnostic compatibility with existing enterprise infrastructure.
    \item \textbf{Programming language flexibility}: Support multiple languages (C++, Python, C\#, NodeJS, etc.) to facilitate integration with diverse services, particularly those built on the Robot Operating System (ROS) \cite{Macenski2022RobotWild}.
    \item \textbf{Multi-layer support}: Ensure robustness across edge, fog, and cloud layers and allow incomplete topologies (\textit{e.g.}, missing a fog layer).
    \item \textbf{Flexible service distribution}: Allow services to be executed on any layer, except for specialized services requiring hardware interactions.
    \item \textbf{Real-time performance}: Select technologies that minimize overhead and meet stringent real-time constraints.
    \item \textbf{Robustness and security}: Provide resilience against connectivity loss and service failures, while ensuring data security both in transit and at rest.
\end{itemize}

While optimized for robotics, the platform is equally applicable to broader Internet of Things (IoT) and distributed system scenarios, including smart cities, smart factories, and more.

\subsection{Contributions}
The contributions of this work are as follows:

\begin{itemize}
    \item \textbf{Service-oriented architecture for distributed robotics}: All computation is performed through services, enabling multi-layer deployment and configuration.
    \item \textbf{Dynamic data transfer}: Services dynamically share and consume data, with the platform automatically connecting related services across heterogeneous brokers.
    \item \textbf{Configurable services}: Standardized configurations with customizable options for deployment-specific needs, ensuring that services can adapt to the execution environment.
    \item \textbf{Monitoring solution}: Integrated mechanisms for performance metrics collection, enabling optimization and system health monitoring at the layer, node, and service level.
    \item \textbf{ROS integration}: Seamless integration of ROS packages into the platform to cover most of the robotics use cases \cite{Portugal2024InquiringMiddleware}, while maintaining service modularity.
    \item \textbf{Multi-layer deployment}: Demonstration of edge, fog, and cloud integration in a multi-cluster/multi-tenant environment, showing how services can be strategically allocated to balance latency, resource constraints, and computation speed.
\end{itemize}

\subsection{Organization}
This paper is organized as follows. The state of the art is discussed in Section \ref{section:related_work}. Sections \ref{section:proposed_solution} and \ref{sec:blazeflow} detail the architecture of the proposed BlazeAIoT platform and its alignment with the objectives outlined in Section \ref{section:motivation}. Section \ref{section:evaluation} describes experimental infrastructure, including the robot, the various ROS packages, and the network topology. Section \ref{section:results} presents the evaluation scenarios and analyzes the platform performance across multiple layers. Section \ref{section:discussion} presents the discussion of the results and compares the platform with the objectives. Section \ref{section:conclusion} concludes with a discussion of results and outlines directions for future work.

\section{Related Work}\label{section:related_work}
Many concepts in distributed robotics are not unique to robotics. Similar principles are applied in the Internet of Things (IoT), which also relies on edge, fog, and cloud topologies \cite{Ahmed2023TheComputing}. The authors describe two common IoT architectures: the three-layered architecture (perception, network, and application layers) and the service-oriented architecture (sensing, network, service, and interface layers). In \cite{Gangrade2025EdgeSystems}, the authors discuss the use of edge and fog computing in cyber-physical systems (CPS) to address limitations of cloud computing, such as high latency, limited bandwidth, and security/privacy concerns. Robots can be considered CPS because they are controllable systems that require real-time interactions.

Cloud robotics has been extensively studied over the past decade. In \cite{Hu2012CloudApplications}, the authors present the cloud robotics concept and detail the implementation challenges. They identify computational challenges that can generate a high volume of data for network transfer, along with communication challenges such as network latency and potential connectivity loss. The XBot2D project \cite{Muratore2023XBot2D:Robotics} proposes a cloud robotics architecture for a Robot-as-a-Service (RaaS) scenario. To mitigate network faults, the XBot2D platform switches robot processes locally when network degradation is detected. The task is executed in both environments (cloud and local) at different rates (200 Hz and 1 KHz) to match resource availability. However, this approach continuously consumes resources across both layers, which is problematic for battery-powered robots.

Fog and cloud robotics solutions are compared in \cite{KrishnaChandGudi2018FogInteraction} for human-robot interaction scenarios. The authors propose an offloading methodology to optimize the task execution. If the robot is unable to execute the task, it offloads it to the fog computing, which then offloads it to the cloud as a last resort. While cloud computing offers virtually unlimited resources, the multi-layer offloading introduces delays that can hinder real-time execution. Authors in \cite{Khalid2019IntelligentE-polling} also introduce fog computing to an E-poll solution to reduce latency and improve the reliability of cloud computing. They also propose a configuration solution for the nodes that enables them to fetch their own configuration upon connecting to the network. The received configuration will influence packet routing to optimize communication speed and reliability. Fog computing has also been explored in other domains, such as healthcare \cite{Aiswarya2021AComputing}. Authors in \cite{Ksentini2020HowCities} use a resource management simulator to prove the benefit of fog computing compared with cloud computing for a real-time and delay-sensitive scenario.

Edge robotics, in conjunction with cloud robotics, can benefit from 5G networks, as introduced in \cite{Roy2025QualityRobotics}. The authors proposed a quality-of-control-based resource allocation solution to optimize computing placement with respect to network delay, reliability, and task periodicity. Although promising, the placement strategy is static and does not account for dynamic changes in the robot environment. This aspect is not part of the actual work, but it is considered in the architecture for future improvements. Similarly, \cite{Baxi2025Achieving5G} explores 5G-enabled edge robotics in a pick-and-place process. A hybrid fallback mechanism is introduced that combines Quality of Service (QoS)-aware robot adaptation to address network instability and mitigate robot constraints, with Control-aware Dynamic QoS orchestration to ensure task execution without duplicating processes across layers. They employ a neural network in their solution and emphasize the need for a robust simulator to enable integration in real-world settings.

Distributed robotics is frequently motivated by the need to offload Artificial Intelligence (AI)-related tasks to reduce the robot's computational burden. In that regard, the addition of a digital twin, as proposed in \cite{Riedlinger2022ConceptTwins}, is a valuable approach for collecting training data and leveraging simulation to improve AI models. The authors propose a distributed framework using Kafka as a data broker and ROS for robot communication. They divide the application into three functional layers: AI inference, the shop floor (with physical devices), and the edge cloud (responsible for the digital twin and non-AI computation). Orchestration of an AI model within a distributed robotics solution can also be challenging. The authors in \cite{Suneel2025EdgeMethods} present a load-balancing solution for AI model inference that ensures real-time execution while prioritizing energy efficiency, particularly on battery-powered edge nodes.

Another essential aspect of distributed robotics is the ability to share data efficiently and reliably across layers. Authors in \cite{Serdaroglu2022EffectSystem} propose a comparison of the message-queue-based communication from Kafka with a conventional HTTP-based communication. They demonstrate that using a broker reduces latency as the number of nodes increases. Message volume significantly affects network latency. Authors in \cite{Montella2018PerformanceApproach} present an HTTP-based communication that segments a message optimally to be sent on multiple concurrent threads. Each segment is encrypted, and the server reassembles the message, thereby providing a secure means of sending such messages and reducing overall latency. The platform solution proposed in \cite{Sakai2022IoTCloud} also introduces a distributed architecture with a device registration process that changes the solution's topology, using a flow-sharing method to enable data flow. The authors implemented data communication using a Remote Procedure Calls (RPC) based approach in the cloud and used Message Queuing Telemetry Transport (MQTT) for inter-layer communication. MQTT is limited to high-volume, high-throughput scenarios, as explained in \cite{Melancon2023BlazeFlow:Applications}, which can be problematic for robotics applications. Another approach to minimize the latency is to use data compression to reduce the message size. Authors of \cite{Qingqing2019VisualNetwork} propose an analysis of the compression effect for a visual odometry scenario. The compression shows a slight degradation in odometry precision while significantly improving bandwidth savings. A balance between model precision and performance improvement can be achieved, depending on the task's criticality.

The aforementioned works highlight the diversity of approaches for distributed robotics and IoT communication, ranging from cloud‑centric architectures to fog and edge solutions, often relying on specific brokers or static placement strategies. Our previous work, BlazeFlow \cite{Melancon2023BlazeFlow:Applications}, introduced an initial vision of multi‑layer data bridging and demonstrated the potential of dynamic communication across heterogeneous brokers. While BlazeFlow laid the foundation, it did not yet provide a complete platform for service deployment and orchestration. In contrast with the aforementioned approaches, BlazeAIoT advances this vision by adopting a service‑oriented architecture that integrates configuration, monitoring, and adaptive data distribution across edge, fog, and cloud layers. This positions BlazeAIoT as a unified and resilient platform for real‑time distributed robotics.

\section{Proposed Solution}\label{section:proposed_solution}
This section outlines the architecture of the proposed platform, named BlazeAIoT, which supports an autonomous robotics solution. We introduce and motivate each subsystem to meet the objectives.

\subsection{System Overview}
\begin{figure*}
    \centering
    \includegraphics[scale=0.38]{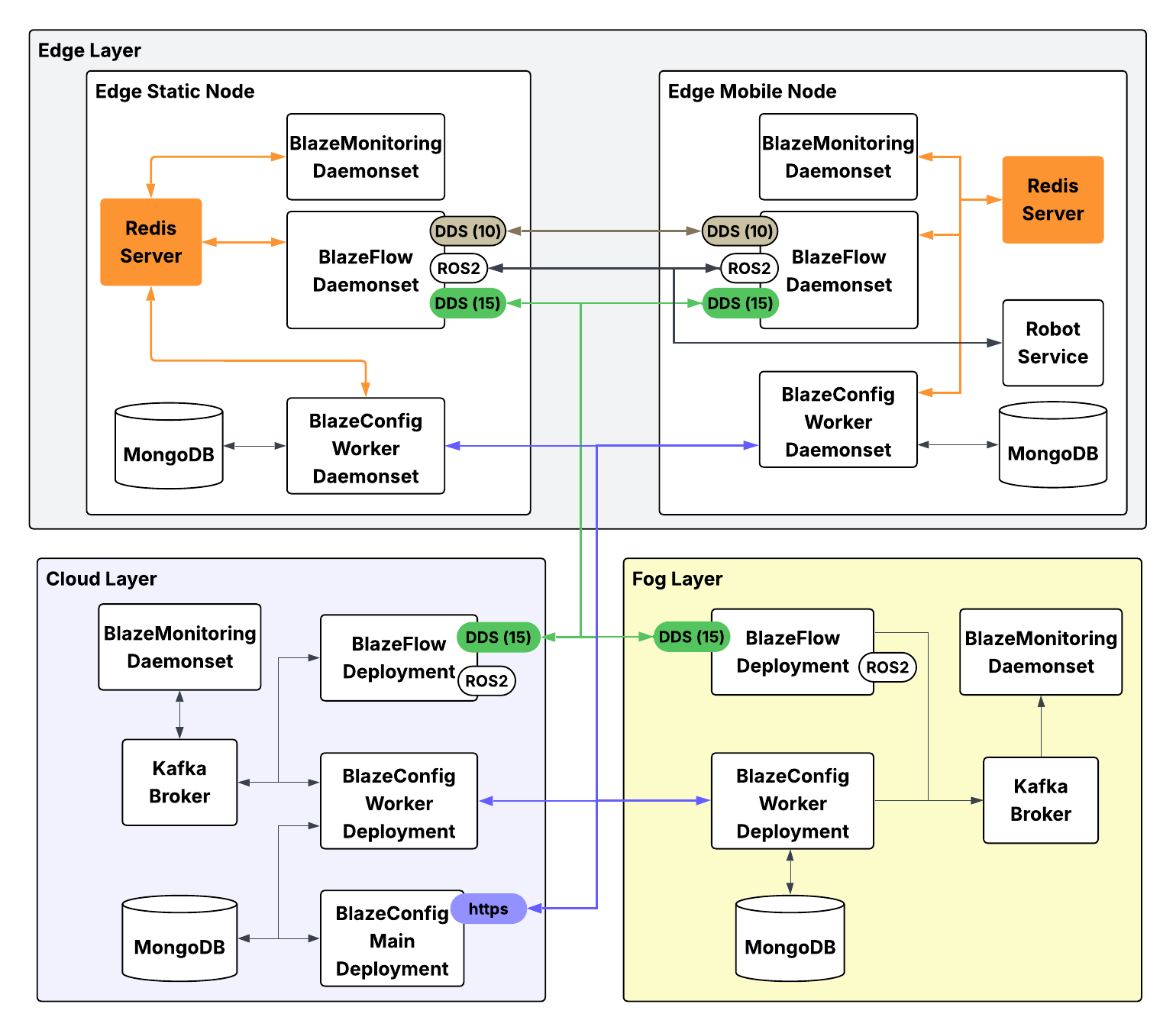}
    \caption{BlazeAIoT High-level architecture.\label{fig:architecture}}
\end{figure*}
The high-level architecture of BlazeAIoT is represented in Fig. \ref{fig:architecture}. The system is divided into three distinct subsystems: edge, fog, and cloud layers. All layers are interconnected via the Data Distribution Service (DDS) as the inter-layer communication protocol, thereby avoiding the need to always rely on the fog layer to forward messages, which would introduce more latency. Additionally, it enables a more flexible topology in which each layer may be optional or replicated for redundancy, ensuring high availability. Each layer is a Kubernetes cluster managed with the open-source solution Rancher\footnote{https://ranchermanager.docs.rancher.com/}. The edge layer cluster uses the lightweight open-source Kubernetes distribution K3S\footnote{https://docs.k3s.io/}, primarily because it is compatible with the ARM64 architecture and the limited resources of edge nodes. The fog and cloud clusters are running the open-source RKE2 Kubernetes\footnote{https://docs.rke2.io/}.

\subsection{BlazeAIoT SDK: A standard interface for every service}
The BlazeAIoT SDK is a library available in multiple languages. As of now, the supported languages are C++, C\#, and Python, which cover most of the possible scenarios for the autonomous robot solution, but support for other languages could be easily integrated.

The SDK is a utility for integrating with the BlazeAIoT platform. A \texttt{ServiceManager} singleton is created for the service as a central access point for the various managing tools:

\begin{itemize}
    \item \texttt{BaseConfiguration}: Interprets the configuration file and provides access to layer, node, and service configurations.
    \item \texttt{Broker}: Provides generic access to the local broker, enabling the publication of messages on a topic and the subscription to topics with a callback function for event-based scenarios.
    \item \texttt{ConfigManager}: Enables a service to respond to configuration changes by providing a callback function for altering node, layer, and service configurations.
    \item \texttt{FlowManager}: Enables the service to inform other services about the topics it generates (advertises) and the topics it consumes (requests).
    \item \texttt{MonitorManager}: Serves as a wrapper over Prometheus\footnote{https://prometheus.io/}, which is an open source metrics and monitoring middleware. It creates a metrics server that can be exposed and provides an asynchronous method for storing metrics, such as counters and gauges.
    \item \texttt{CacheManager}: Enables some functionalities leveraging Redis\footnote{https://redis.io/community/} as a cache. It is used to share service availability via a watchdog. It creates a timer that refreshes a hash value in the database. This value has an expiration, and is eventually removed from the database if it is not updated within the specified time.
\end{itemize}

The primary reason for using Redis to validate service availability is to synchronize interdependent platform services (for example, BlazeConfig requires BlazeFlow, and BlazeFlow requires BlazeConfig). Otherwise, undefined behavior can occur if the sequence is not respected.

\subsection{BlazeConfig: Multi-layer Configuration}
The BlazeConfig service is a new addition to the BlazeAIoT solution. It enables the management and synchronization of layer, node, and service configurations across all layers of the solution. BlazeConfig is composed of three components:

\begin{itemize}
    \item \textbf{MongoDB Database}: this component is responsible for storing the configuration. At the edge, each node has its own MongoDB\footnote{https://www.mongodb.com/community/} instance, and in the fog and cloud, it can be deployed as a Kubernetes application or on a specific node.
    \item \textbf{BlazeConfig Main Service}: this service is hosted on the most centralized layer (the cloud layer when available, otherwise the fog layer). It exposes a REST API to manage globally the configuration of every aspect of the system. All the data is stored in the MongoDB database.
    \item \textbf{BlazeConfig Worker Service}: this service manages the local configuration for the node/layer. It is deployed as a Kubernetes DaemonSet on the edge layer (each node runs a single instance connected to the local MongoDB database) and as a Kubernetes Deployment on the fog and cloud layers. Each service gets its configuration from this service through the local broker (intra-node for the edge and intra-layer for the fog and cloud).
\end{itemize}

The worker service gets the latest configuration for the whole layer directly from the main service. The received configuration is used to synchronize the worker MongoDB database. After that, all interactions between the main service and the brokers are handled by the local broker. Each service affected by the change receives a notification via the local broker and takes appropriate action. The action can be to update the service's local parameters or the node/layer parameters, such as the database IP address. Each service is responsible for implementing its actions in response to a configuration change.

\subsection{BlazeFlow: Multi-layer communication}
A preliminary vision of BlazeFlow was first introduced in a workshop publication \cite{Melancon2023BlazeFlow:Applications}. The goal was to enable efficient, scalable multi-layer communication in distributed systems. The middleware introduced three different communication brokers. The intra-node communication uses Redis for local communication within a node. Intra-layer communication varied across layers. For node-to-node communication within the edge layer, DDS middleware was used; the fog and cloud layers used Kafka\footnote{https://kafka.apache.org/}. Finally, inter-layer communication, used to transfer messages between layers, was implemented using the DDS middleware.

BlazeFlow can also be leveraged for interoperability by enabling broker plugins that support additional protocols, such as ROS2. BlazeFlow bridges supported protocols in any direction (\textit{e.g.} ROS2 $\leftrightarrow$ DDS $\leftrightarrow$ Kafka).

While retaining the functionality name from previous work, BlazeFlow was significantly extended to integrate with BlazeAIoT by adding the capability to perform dynamic data distribution, defining the network configuration to be used in Kubernetes, enabling communication between multiple clusters, and introducing the concept of adaptive data distribution. Therefore, it represents a significant novel contribution to this work.

\subsection{BlazeMonitoring: System Monitoring}
BlazeMonitoring is a service that extracts essential platform metrics and makes them available for future optimization. The service is divided into two separate applications. 

The BlazeMonitoring DaemonSet is spawned on every node for every layer. It is responsible for collecting node metrics, including CPU, GPU, memory, temperature, and power (when available). It is also used to monitor the latency between nodes in the solution. A ping timer task is executed at a configurable period of time (60 seconds) and sends a ping to all the nodes inside the solution (even the nodes outside the layer). When a node receives a ping message, it replies with a pong message, and the elapsed time divided by two represents the average latency between the two nodes. The primary reason for creating this task, rather than a typical Linux ping, is that it also accounts for the latency introduced by BlazeFlow bridges. 

The BlazeMonitoring Deployment can be used to collect specific metrics from a particular node. For this work, the message latency is computed to assess the impact of message size. Each message is timestamped before transmission by the first service. The monitor service listens to preconfigured topics and computes the elapsed time, which represents message latency (compared with the network latency from the DaemonSet).

All metrics are stored in the Prometheus server via the BlazeAIoT SDK and can be used to create Grafana dashboards or extracted for manipulation in a notebook, as used for the experimental portion of the project.

\section{BlazeFlow as a Dynamic Data Bridge Solution}\label{sec:blazeflow}
This section discusses the key components integrated into BlazeFlow to meet the objectives of BlazeAIoT.

\subsection{Dynamic Data Distribution}
BlazeFlow provides a dynamic data distribution paradigm that supports autonomous robotics solutions. This concept enables dynamic service spawning, ensuring that services receive the expected messages and that other services can process them.

BlazeFlow serves as a dynamic bridge that supports multi-layer communication. When a service is spawned on a node, it provides the advertised and requested topics to the local BlazeFlow service (via intra-node communication on edge nodes and intra-layer communication on fog/cloud nodes). BlazeFlow takes three actions from this information:

\begin{itemize}
    \item Store the advertise/request topic to be able to take action when another service is involved with the same topic.
    \item Create a bridge between two protocols if a pair of advertise/request exists on the layer.
    \item Forward the advertise/request to the other layers. Source validation (node and layer) is performed to prevent message loops.
\end{itemize}

\begin{figure*}
    \centering
    \includegraphics[scale=0.45]{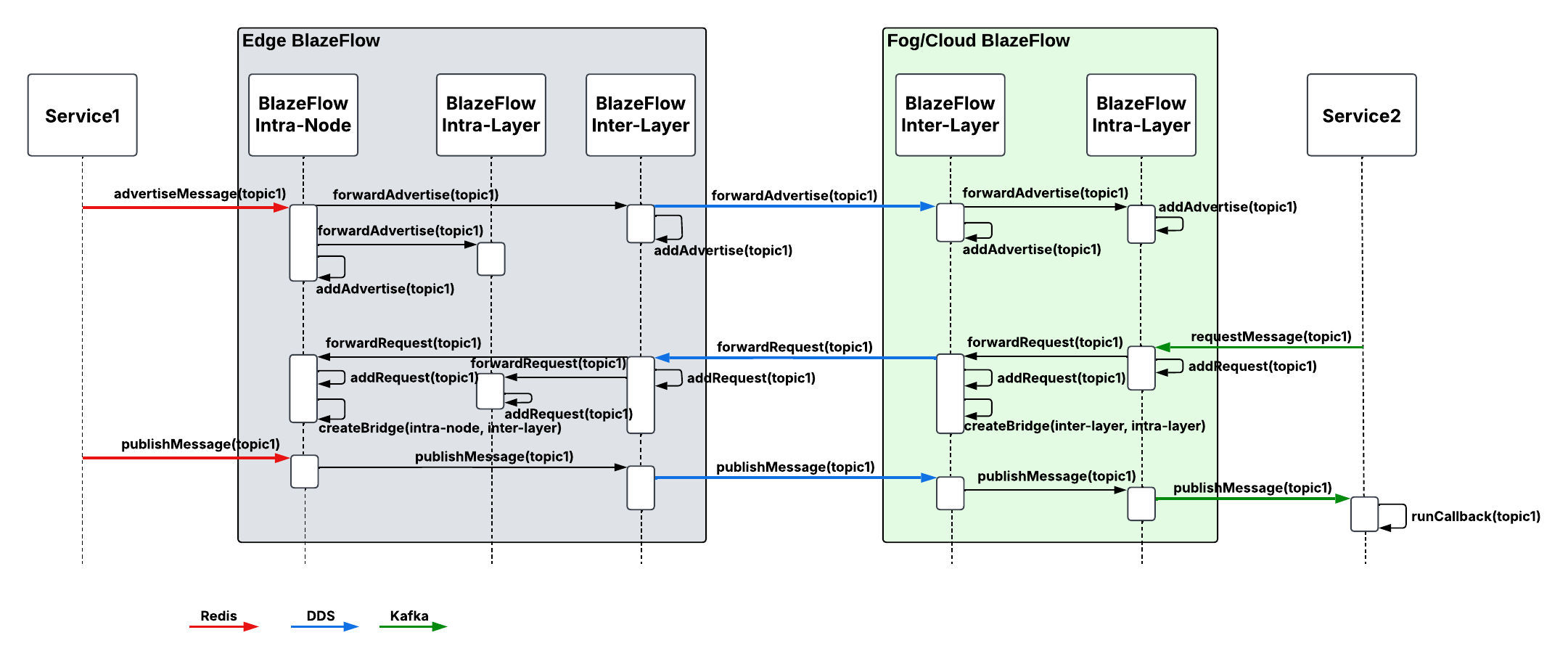}
    \caption{Multi-layer handshaking sequence.\label{fig:seq_bf}}
\end{figure*}

\subsubsection{Bridge Creation Example}
An example sequence for creating a bridge and the flow of a bridge message is shown in Fig. \ref{fig:seq_bf}. The two BlazeFlow boxes include the various brokers that they support. Each broker instance is a subscriber to the remote BlazeFlow brokers. Each received message triggers a task to perform the BlazeFlow logic. In this example, Service1, hosted on an edge node, must send data to Service2, which is hosted on either the fog or the cloud layer (which makes no difference to the logic). The first action is for Service1 to announce the advertised topic 'topic1', given that \texttt{Service1} wants to publish on that topic. The message is sent to the intra-node broker and received by the BlazeFlow service. The advertised message is forwarded to the intra-layer and inter-layer brokers and stored for each broker. At the next layer, the BlazeFlow service receives the advertised message via the inter-layer broker and forwards it to the intra-layer broker. It then stores the advertised topic for each broker.

The following action is for Service2 to announce the requested topic 'topic1' because Service2 is interested in receiving messages for that topic. The message is sent to the intra-layer broker and accepted by the BlazeFlow service. The requested message is forwarded to the inter-layer broker and stored for each broker. At the edge layer, the inter-layer broker receives the requested message and forwards it to the intra-layer and intra-node brokers, where each node stores it. 

Since a pair of advertise/request is found, the BlazeFlow service needs to create bridges. On the edge node, a bridge is made to transfer data from the intra-node broker to the inter-layer broker. On the fog/cloud layer, a bridge is created to transfer data from the inter-layer broker to the intra-layer broker.

The final action is an example of a message sent by Service1 on the topic 'topic1'. The BlazeFlow bridge receives the message and automatically publishes it to the inter-layer broker. On the other layer, the BlazeFlow bridge gets the message from the inter-layer broker and forwards it to the intra-layer broker. The Service2 receives the message and invokes the action associated with the message to perform a task (\textit{i.e.}, the callback).

The logic of bridge creation depends on the availability of an advertise/request pair for the same topic and its origin. The goal of BlazeFlow is to transparently establish a continuous data flow from the original service to the destination services. If the destination is within the same layer, the intra-layer communication mechanism is used (simple case), and no bridging is required; however, if the destination is outside the layer, inter-layer communication is used. Multiple destinations may be necessary, requiring BlazeFlow to route messages across various brokers.

\begin{figure}
    \centering
    \includegraphics[scale=0.5]{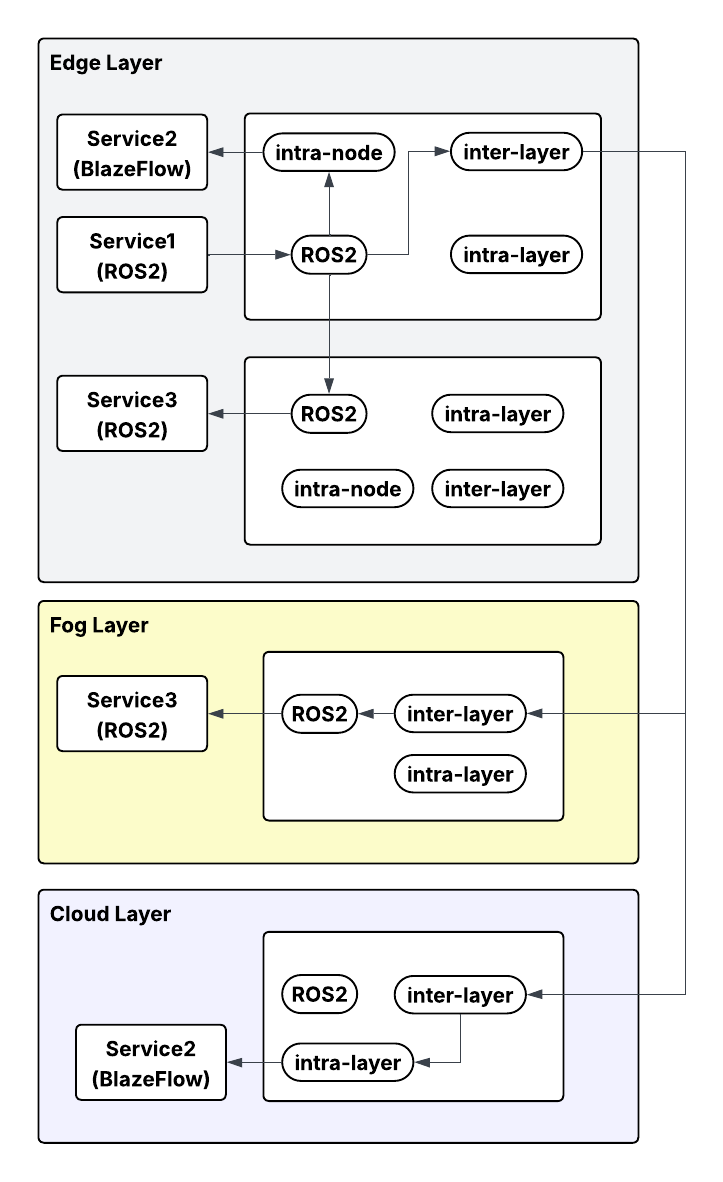}
    \caption{Multi-bridge scenario example.\label{fig:bf_multi_broker}}
\end{figure}

\subsubsection{One-to-many Scenario Example}
Fig. \ref{fig:bf_multi_broker} represents an example of a complex scenario with one publisher (Service1) and multiple subscribers (Service2 and Service3) dispersed on various layers. A flow diagram is shown in this example to abstract away some details. Service1 provides data via the ROS2 broker. 

Service2 is deployed on the same node and on the cloud layer. Service2 is using the default BlazeFlow broker. To receive data at the edge layer, a bridge is created to publish ROS2 messages to the intra-node broker (Redis Pub/Sub). For the Service2 hosted in the cloud layer, a bridge is required to transmit ROS2 data to the inter-layer broker (DDS). In the cloud layer's BlazeFlow, a bridge is necessary to transfer data from inter-layer communication (DDS) to intra-layer communication (Kafka), enabling Service2 to receive it.

Service3 is deployed on a separate edge node and on the fog layer. Service3 is expecting data from the ROS2 broker. Since, as explained earlier, ROS2 is considered local within the same layer, the Service3 hosted on a separate edge node will receive data directly from the same ROS2 network without requiring a bridge. For the Service3 hosted on the fog layer, it will utilize the existing bridge on the edge layer (ROS2 $\rightarrow$ inter-layer). Additionally, a bridge will be created on the fog layer's BlazeFlow to transfer data from the inter-layer broker (DDS) to the ROS2 layer.

\subsection{Network Considerations}
Brokers that are not directly provided by the platform (intra-node, intra-layer, and inter-layer) are considered accessible from within the layer because all nodes are part of the same Kubernetes cluster. A macvlan interface is created on the cluster to share the data using Multus\footnote{https://github.com/k8snetworkplumbingwg/multus-cni} and Whereabouts\footnote{https://github.com/k8snetworkplumbingwg/whereabouts}. This ensures that all nodes are sharing the same virtual network.

Inter-layer communication is enabled by Submariner\footnote{https://submariner.io/}, which establishes an IPsec tunnel between clusters. Other, more efficient networking solutions are available but are more complex to configure; therefore, the default configuration was used. IPsec tunneling is highly CPU-intensive, which limits network performance; however, it ensures that inter-layer traffic is encrypted. Due to the limited resources, the available bandwidth for inter-layer communication is about 160 Mbps. This could be improved for industrial deployment by using gateway nodes with a faster CPU, or by using a different cable driver technology.

\subsection{Adaptive Data Distribution}
To optimize data flow, all bridge actions are performed in separate tasks submitted to a thread pool sized according to the node's available resources. The same applies to the service callbacks. The cost of this parallelization is increased resource consumption (memory and compute), but it also enables better performance to meet real-time constraints. Without this parallelization, the support for large messages significantly slowed the overall data distribution, making it unusable in a distributed robotics scenario.

To support large messages, such as camera images, a rate-limiter logic was implemented in BlazeFlow. Small messages are prioritized, and large messages have their rate reduced by skipping frames to match the allowed rate within the bandwidth budget. For instance, if we were to replace IPsec with an unencrypted tunnel (at the cost of reduced data privacy), the allowed bandwidth budget would increase, and the framework would automatically adapt its rate-limiting logic.

The rate-limiting logic implemented by BlazeAIoT is a two-tier hierarchical algorithm that combines client-level bandwidth allocation with publisher-level token-bucket regulation. At the client level, available bandwidth is defined in Eq. \ref{eq:bandwidth}. 

\begin{equation}\label{eq:bandwidth}
B_{\text{avail}} = L_{\text{client}} \cdot \frac{1024^2}{8} bytes/sec
\end{equation}
where $L_\text{client}$ is the limit in Mbps.

Bandwidth is distributed among publishers using a priority-based allocation strategy: standard publishers receive their advertised rates first. In contrast, large-message publishers are allocated the remaining bandwidth, with a minimum threshold of $r_{\min} = 2$ Hz to prevent starvation. The algorithm accounts for message overhead with a factor $\alpha = 1.02$ and enforces a configurable per-message maximum bandwidth percentage $\beta = 0.95$ (default 95\%) to prevent monopolization. Publishers are sorted by bandwidth demand $D_i = r_i^{\text{adv}} \cdot s_i \cdot \alpha$ in descending order, where $r_i^{\text{adv}}$ is the advertised rate and $s_i$ is the message size in bytes. For each publisher $i$, the allocated rate is computed with Eq. \ref{eq:a_rate}.

\begin{equation}\label{eq:a_rate}
r_i^{\text{alloc}} = \min\left(r_i^{\text{adv}}, \frac{B_{\text{remaining}}}{s_i \cdot \alpha}, \frac{\beta \cdot B_{\text{avail}}}{s_i \cdot \alpha}\right)
\end{equation}
where $B_{\text{remaining}}$ is updated after each allocation with Eq. \ref{eq:b_remain}. 

\begin{equation}\label{eq:b_remain}
B_{\text{remaining}} \leftarrow B_{\text{remaining}} - r_i^{\text{alloc}} \cdot s_i \cdot \alpha
\end{equation}

At the publisher level, each topic employs a token bucket filter with capacity $C = 2$ tokens, as defined in Eq. \ref{eq:token}.

\begin{equation}\label{eq:token}
T(t) = \min\left(C, T(t - \Delta t) + r_i^{\text{alloc}} \cdot \Delta t\right)
\end{equation}
where $T(t)$ is the token count at time $t$ and $\Delta t$ is the elapsed time in seconds. 

A message is transmitted only if $T(t) \geq 1$, consuming one token per publish: $T(t) \leftarrow T(t) - 1$. This rate-limiting mechanism allows brief catch-up bursts above the nominal rate while maintaining long-term average compliance. It focuses on delivering small-payload messages at their publish rates, as much as possible, while reducing the rate of large-payload messages, which consume most of the bandwidth.

To reduce message size, the payload can be compressed using the LZ4 algorithm with a configurable compression level. The default level for large messages is 10, which provides a reduction of 40-60\% depending on the image content. Since a compressed image has not a constant size, the rate-limiting algorithm always consider the maximum size for a given topic and will adjust the rate when a bigger message is received.

\section{Evaluation Testbed}\label{section:evaluation}
This section describes the experimental environment used to evaluate the BlazeAIoT platform. It presents the infrastructure topology, including the edge, fog, and cloud layers, and describes the robot and the various components needed for the evaluation.

\subsection{Infrastructure}
\begin{figure*}
    \centering
    \includegraphics[scale=0.37]{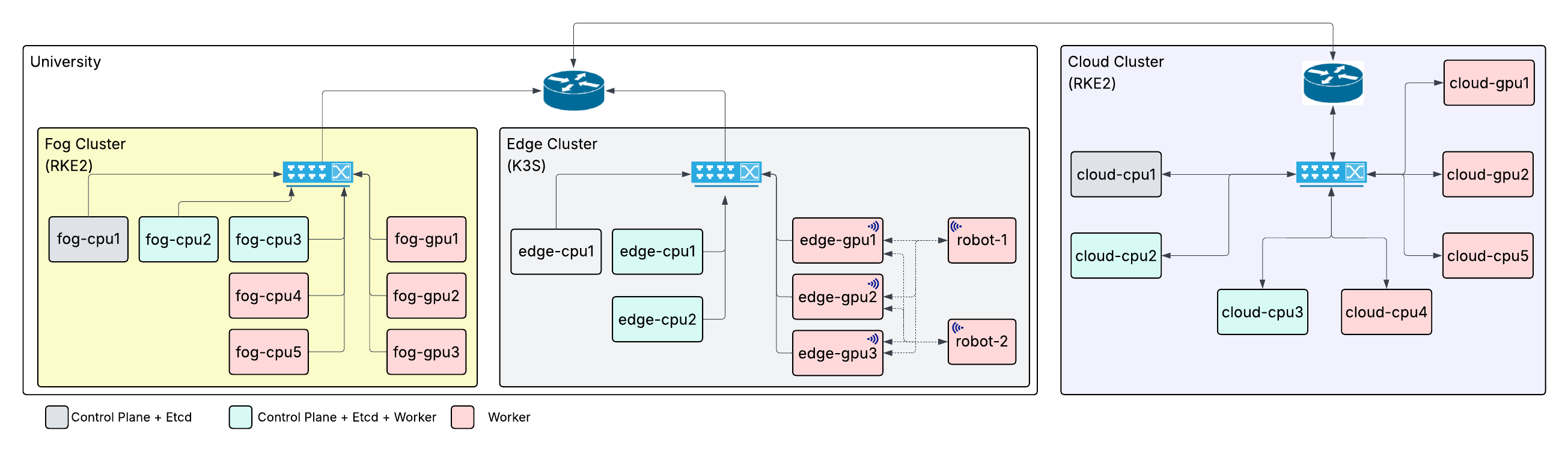}
    \caption{Experimental server infrastructure.\label{fig:infra}}
\end{figure*}

To validate the solution, the infrastructure detailed in Fig. \ref{fig:infra} is used. Due to network restrictions, a separate server was configured to act as an on-premise cloud server.

All layers are providing a Kubernetes cluster. To facilitate configuration, the open-source solution Rancher was installed. The primary reason for selecting this application is its ability to deploy RKE2 Kubernetes clusters (for the fog and cloud layers) alongside K3S Kubernetes clusters (for the edge layer).

To allow multi-cluster communication, the application Submariner was configured. Each layer provides at least one gateway through which network communication flows. Each cluster's gateway shares a virtual network managed by the main deployment (configured as an operator).

Rancher provides a monitoring application that includes a Prometheus server that gathers metrics to monitor the cluster and Kubernetes performance. It also provides a Grafana deployment with multiple pre-bundled dashboards.

The layer's default configuration, its nodes, and all services are stored in a Kubernetes ConfigMap that can be mounted as a volume in the services. This is useful when the BlazeConfig service is unavailable, enabling configuration of basic functionality, such as the broker, database, and cache addresses and ports.

\subsubsection{Cloud cluster}
All the nodes in the cloud cluster are connected to the same network switch. The switch has access to the internet and the other clusters through a router. Since the cloud network is located near the other clusters, the scenario is made more realistic by configuring the router to introduce latency on the port connected to the router hosting the fog and edge clusters. A 50 ms latency with 10 ms jitter is added along with a packet loss of 0.01\%\cite{MicrosoftAzureLearn}.

The cloud cluster is an RKE2 cluster composed of seven nodes. Since using real servers with virtualization was not possible, each node is a computer with a different configuration. Table \ref{tab:cloud_cluster} describes the configuration of every node.

\begin{table}[!ht]
    \centering
    \caption{Cloud cluster nodes}
    \begin{tabular}{|l|c|c|c|l|c|}
        \hline
        \textbf{Node name} & \textbf{CPU} & \textbf{RAM} & \textbf{GPU (Memory)} \\
        \hline
        cloud-cpu1 & 2 & 16 Gb & 0 \\
        \hline
        cloud-cpu2 & 2 & 16 Gb & 0 \\
        \hline
        cloud-cpu3 & 4 & 16 Gb & 0 \\
        \hline
        cloud-cpu4 & 4 & 32 Gb & 0 \\
        \hline
        cloud-cpu5 & 4 & 32 Gb & 0 \\
        \hline
        cloud-gpu1 & 12 & 32 Gb & 1 (8 Gb) \\
        \hline
        cloud-gpu2 & 20 & 32 Gb & 1 (24 Gb) \\
        \hline
        \hline
        Total & 48 & 176 Gb & 2 (32 Gb) \\
        \hline
    \end{tabular}    
    \label{tab:cloud_cluster}
\end{table}

A Kafka controller/broker application is deployed to serve as intra-layer communication. Inter-layer communication using DDS is enabled through a Kubernetes service exposed via Submariner. A Redis application is also deployed on the cluster.

The cloud-cpu3 has MongoDB installed outside Kubernetes and exposed its service for the other nodes.

The BlazeConfig Main service is deployed on the cloud cluster, and the REST API is exposed through a Kubernetes service accessible via Submariner. The BlazeMonitoring, BlazeFlow, and BlazeConfig worker applications are also deployed on the cluster.

\subsubsection{Fog cluster}
The fog cluster is also an RKE2 cluster and is very similar in configuration to the cloud one. However, instead of having its own network, it shares the edge cluster's network via the same router. All nodes are connected to the same network switch, which is connected to the router that provides internet access and connectivity to the cloud cluster network. The Table \ref{tab:fog_cluster} details the configuration of every node in the fog network.

\begin{table}[!ht]
    \centering
    \caption{Fog cluster nodes}
    \begin{tabular}{|l|c|c|c|l|c|}
        \hline
        \textbf{Node name} & \textbf{CPU} & \textbf{RAM} & \textbf{GPU (Memory)} \\
        \hline
        fog-cpu1 & 8 & 8 Gb & N/A \\
        \hline
        fog-cpu2 & 8 & 16 Gb & N/A \\
        \hline
        fog-cpu3 & 8 & 16 Gb & N/A \\
        \hline
        fog-cpu4 & 2 & 4 Gb & N/A \\
        \hline
        fog-cpu5 & 8 & 16 Gb & N/A \\
        \hline
        fog-gpu1 & 8 & 16 Gb & 1 (8 Gb) \\
        \hline
        fog-gpu2 & 8 & 16 Gb & 1 (8 Gb) \\
        \hline
        fog-gpu3 & 12 & 85 Gb & 1 (48 Gb) \\
        \hline
        \hline
        Total & 62 & 177 Gb & 3 (64 Gb) \\
        \hline
    \end{tabular}    
    \label{tab:fog_cluster}
\end{table}

Similar to the cloud cluster, the fog cluster is configured with Redis and Kafka for intra-layer communication, an exposed service in Submariner for inter-layer communication, and the BlazeAIoT applications (BlazeFlow, BlazeConfig worker, and BlazeMonitoring).

\subsubsection{Edge cluster}
The edge cluster is a K3S cluster because of its node type. The nodes used for this work range from small computers, such as Raspberry Pi 5, to various Jetson nodes (Nano, NX, AGX). Additionally, the robots are equipped with an edge node, specifically a Jetson Orin AGX. The CPU nodes are installed near the network switch to minimize the latency. The GPU nodes, which are Jetsons, are dispersed throughout the test area and connected to the network switch via long Ethernet cables. They also serve as wireless gateways (configured as WiFi access points). The robot is a wireless node that connects to available access points to help minimize latency. The robot nodes also include GPUs, enabling them to perform critical AI tasks.

The Table \ref{tab:edge_cluster} details the configuration of every node in the edge network.

\begin{table}[!ht]
    \centering
    \caption{Edge cluster nodes}
    \begin{tabular}{|l|c|c|c|l|c|}
        \hline
        \textbf{Node} & \textbf{Node} & \textbf{CPU} & \textbf{RAM} & \textbf{GPU}      \\
        \textbf{name} & \textbf{type} &              &              & \textbf{(Memory)} \\  
        \hline
        edge-cpu1 & Static & 4 & 8 Gb & N/A \\
        \hline
        edge-cpu2 & Static & 8 & 16 Gb & N/A \\
        \hline
        edge-cpu3 & Static & 8 & 16 Gb & N/A \\
        \hline
        edge-cpu4 & Static & 16 & 16 Gb & N/A \\
        \hline
        edge-gpu1 & Gateway & 6 & 8 Gb & 1 (8 Gb$^*$) \\
        \hline
        edge-gpu2 & Gateway & 8 & 16 Gb & 1 (16 Gb$^*$) \\
        \hline
        robot-1   & Mobile & 8 & 32 Gb & 1 (32 Gb$^*$) \\
        \hline
        \hline
        Total     & -      & 66 & 144 Gb & 3 (56 Gb$^*$) \\
        \hline
    \end{tabular}
    $^*$ Shared memory
    \label{tab:edge_cluster}
\end{table}

Each edge node has a local installation of MongoDB and Redis (outside Kubernetes). The local Redis server is used for intra-node communication and caching. Intra-layer communication with DDS is facilitated by a Multus/Whereabouts network attachment definition that creates a MACVLAN network shared by all nodes in the cluster. The primary reason for this configuration is that it enables UDP multicast, which can reduce DDS bandwidth requirements when multiple nodes share the same topics. Inter-layer communication is set up the same way as the other clusters, via a Kubernetes service exported by Submariner.

The BlazeFlow, BlazeMonitoring, and BlazeConfig worker applications are deployed as DaemonSets, meaning each node runs one instance.

\subsection{Robot}

\begin{figure}
    \centering
    \includegraphics[scale=0.05]{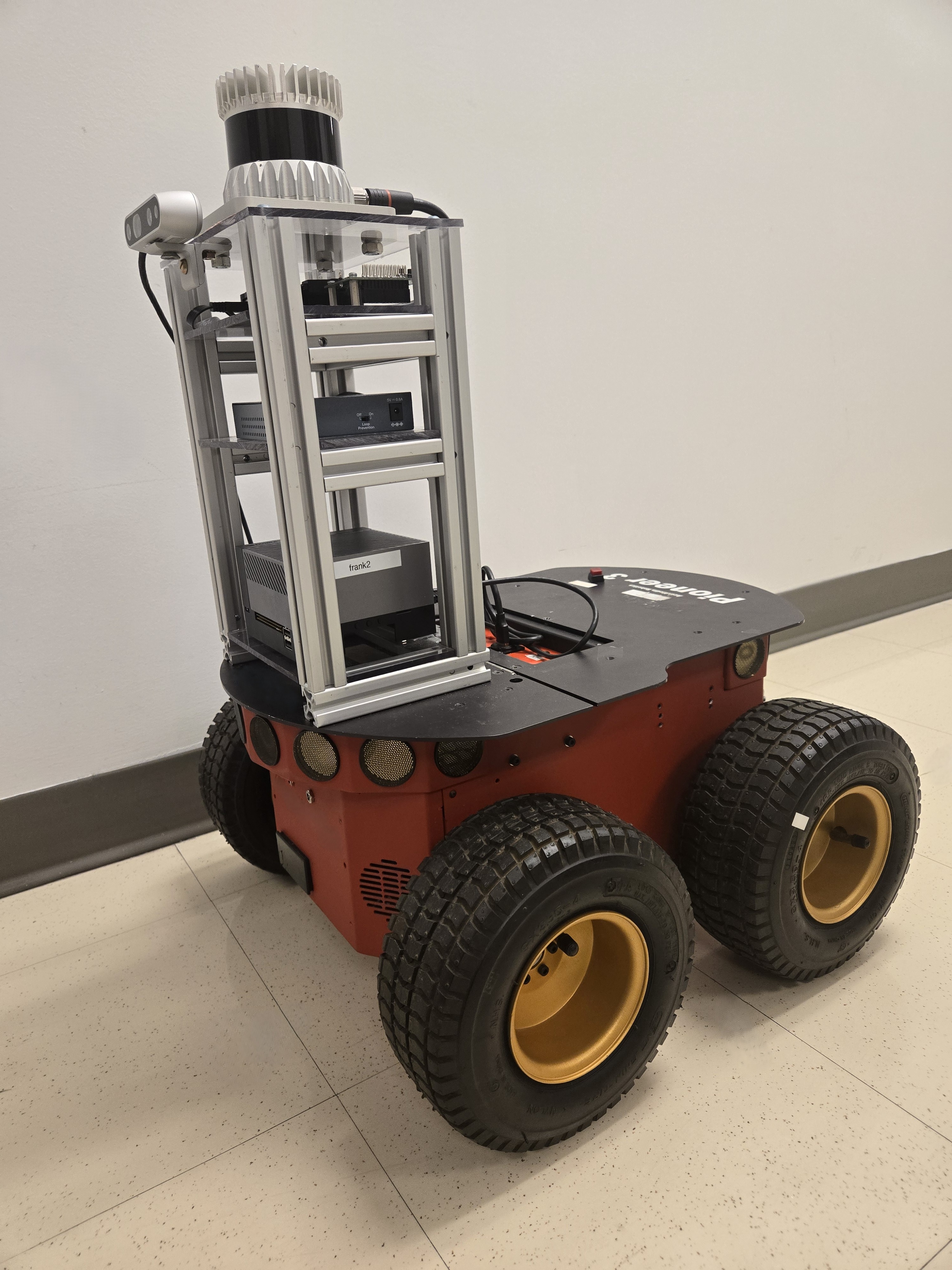}
    \caption{Modified Pioneer 3-AT.\label{fig:robot}}
\end{figure}

The robot is built on the Pioneer 3-AT platform, as described in \cite{Melancon2023TowardsTwin}, with an upgrade from the Jetson computer to the Jetson Orin AGX as represented in Fig. \ref{fig:robot}. The following subsections describe the ROS packages used for the robot navigation.

\subsubsection{Robot Localization Service}
ROS2 uses a set of standard coordinate frames to localize robots on the map. These are the map frame, which represents the global area in which the robot operates, the odom frame, which is placed on the robot's starting location in the map frame and tracks the robot's movement relative to their starting position, and finally the robot's frame, which is connected to the robot's base link and represents the robot's position itself.

This standard frame separation allows separation of concerns when operating the robot platform. The map frame can be used to measure the robot's position within the area of interest. However, given the relatively slow update rates of global localization methods such as SLAM and UWB, which tend to produce discrete jumps, they cannot be used directly for low-level robot control. On the other hand, the odom frame is guaranteed to be continuous and thus suitable for robot control, but it suffers from drift due to odometry sensor errors.

Robot localization is performed using a pair of Kalman filters implemented in ROS2 with the Robot Localization package \cite{Moore2016ASystem}. The local filter generates a transform from the odom frame to the base link using wheel odometry and the IMU's accelerometer and gyroscope. 

The robot pose frequency calculated by the SLAM system is constrained by several factors, including the lidar scan rate and the local Kalman filter's position update rate, resulting in a frequency of 1.5 Hz that reduces navigation reliability. To increase this frequency, a global Kalman filter fuses the same odometry data as the local Kalman filter, along with the SLAM pose, which is more discrete than the odometry and has a higher frequency. This Kalman filter interpolates positions between SLAM poses published by odometry, enabling it to publish a global position at 20 Hz.

To have a comparable ground truth, an Ultra Wideband (UWB) module is used to generate absolute position in the map frame. To filter the UWB's discrete, highly noisy data, an identical set of local and global Kalman filters is used, replacing the SLAM pose with the UWB pose to produce a more reliable UWB ground-truth signal at 20Hz.

\subsubsection{Pure Pursuit Service}

The path-following algorithm used is a pure-pursuit controller, which can be viewed as a carrot-and-stick method \cite{Coulter1992ImplementationAlgorithm}. A line is drawn between the waypoints to build a path. The robot has a lookahead distance forming a circle around it. The controller identifies a goal point by finding the intersection of this circle with the line formed by the waypoints, favoring intersections closer to the next waypoint. The controller then uses the robot's localization data and the target point's position to compute a command for the current time step. This process is repeated at 25 Hz per iteration until the robot reaches the goal point, the last waypoint in the list. The controller computing the command from the localization and target point is a Proportional-Integral-Derivative (PID) controller.

\subsubsection{Emergency Stop Service}
This service is a ROS2 package that detects a halt hand gesture (all fingers extended) and generates an emergency stop.

\begin{figure*}
    \centering
    \includegraphics[scale=0.42]{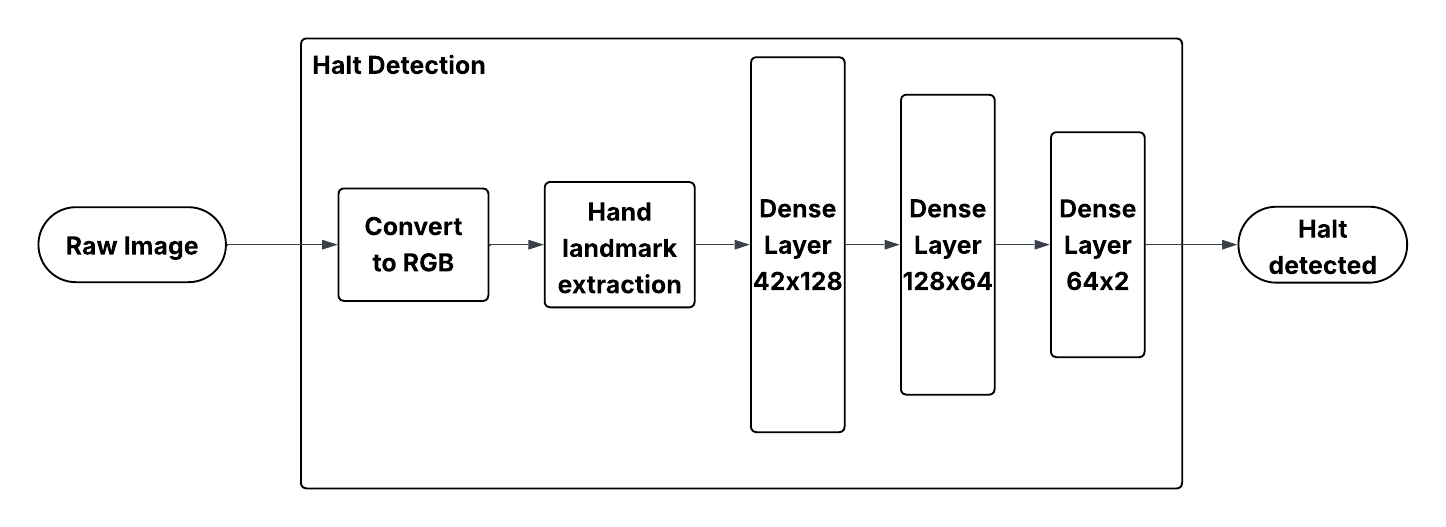}
    \caption{Emergency stop detection model.\label{fig:emergency_stop}}
\end{figure*}

The architecture of the AI model used for the halt detection is detailed in Fig. \ref{fig:emergency_stop}. The hand landmark extraction leverages the MediaPipe hand solution \cite{GoogleLLC2020HandsMediapipe}, which consists of a lightweight convolutional neural network pre-trained on a large dataset of hand images. The landmark detection pipeline passes through three dense layers before making a final decision on whether the halt gesture is detected. The ROS package provides an emergency-stop indication message that the robot can use to halt all motion. The model was trained with the HAGRID dataset \cite{Alexander2024HaGRIDDataset}.

\section{Experimental Results}\label{section:results}
Based on the experimental infrastructure detailed in Section \ref{section:evaluation}, the proposed BlazeAIoT platform is evaluated in two scenarios to assess the impact of computing distribution on a real-time application. Each scenario aims to validate one particular aspect of the platform. The first scenario focuses on small messages with high data rate to assess the feasibility of running a typical navigation at any layer. The second scenario combines large messages with those from the first scenario to evaluate adaptive data distribution and its impact on a critical scenario, such as an emergency stop.

\subsection{Scenario 1: Navigation}
The first scenario is a simple navigation scenario without active obstacle avoidance. The robot navigates the lab using the pure pursuit algorithm, moving from an office room to the end of an open area.

The experiment is divided into five steps. Step 1 is to run all computations locally on the robot edge node, directly in ROS2, rather than via the BlazeAIoT framework. This step is considered the baseline for experimentation. The goal is to assess the platform's overhead and the impact of latency on a typical navigation scenario when running services at any layer by comparing with the baseline execution.

The following steps are performed within the BlazeAIoT framework. Step 2 runs all the robot services on the robot's edge node. Step 3 keeps the P3AT services on the robot's edge node, while all other services run on the edge layer (remote to the robot). Steps 4 and 5 are similar to the third step, but service execution that is not related to the robot itself occurs on the fog and cloud layers, respectively.

\begin{figure*}
    \centering
    \includegraphics[scale=0.42]{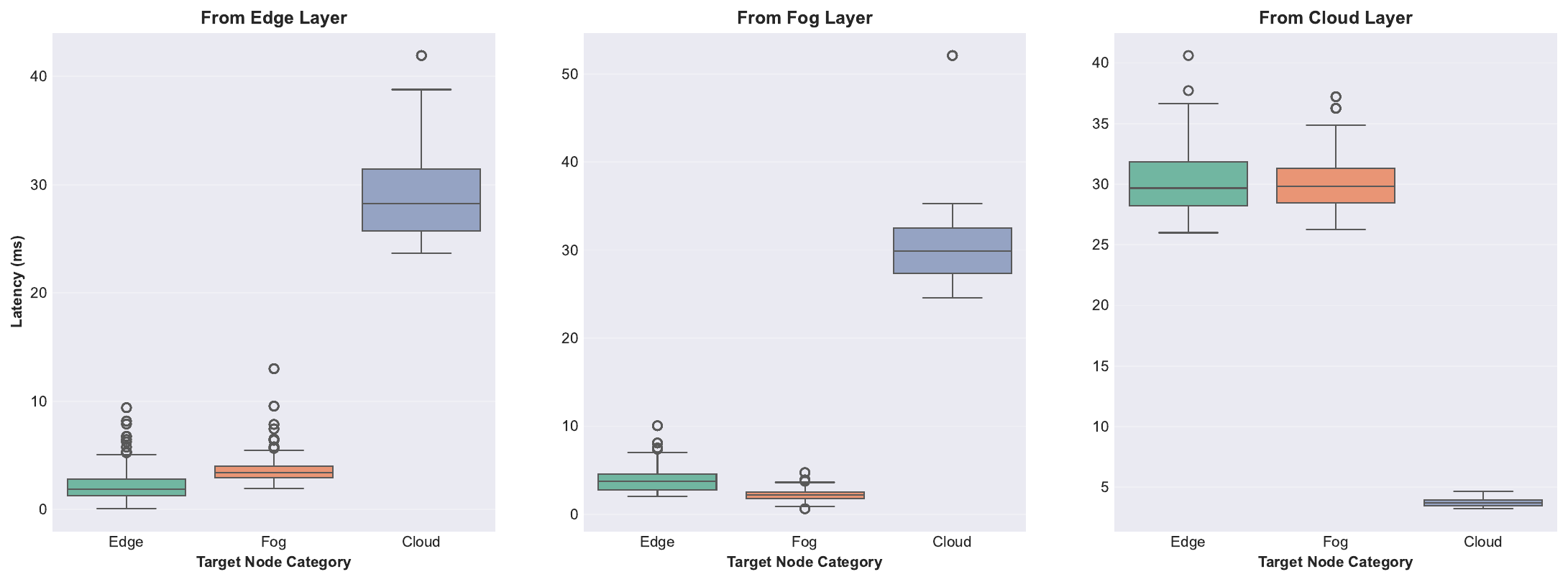}
    \caption{Latency distribution.\label{fig:latency}}
\end{figure*}

To help interpret the results, Fig. \ref{fig:latency} shows the latency distribution observed for each layer when pinged by every other layer using the BlazeMonitor ping service. The latency on the same layer is approximately 2.4 ms in the edge layer, 2.2 ms in the fog layer, and 3.8 ms in the cloud layer. Communication with the cloud is significantly affected by latency/jitter emulation, resulting in an average latency of approximately 30.1 ms. Communication with the fog exhibits an average latency of approximately 3.7 ms from the edge layer.

Table \ref{tab:ros_latency} shows the average latency of the ROS topics along with their rate of production. To achieve this latency, a subscriber probe is created at each layer, and each subscriber computes latency by comparing the header timestamp (generated by the ROS2 node) with the local time. All nodes are synchronized to the same Network Time Protocol (NTP) server, minimizing time drift.

\begin{table*}
    \centering
    \caption{ROS topic latency}
    \label{tab:ros_latency}
    \begin{tabular}{|c|c|c|c|c|c|}
        \hline
        Topic & Rate & Robot        & Edge         & Fog          & Cloud  \\
              & (Hz) & Latency (ms) & Latency (ms) & Latency (ms) & Latency (ms) \\
        \hline
        encoder\_odom & 49.71 & 3.26 & 12.11 & 10.76 & 29.26 \\
        \hline
        odometry/global & 19.09 & 7.68 & 18.98 & 16.22 & 32.81 \\
        \hline
        pose & 1.09 & 218.68 & 285.18 & 219.71 & 216.00 \\
        \hline
        scan & 4.56 & 142.84 & 148.70 & 147.37 & 155.08 \\
        \hline
        imu & 49.17 & 3.30 & 11.26 & 9.38 & 28.14 \\
        \hline
        /tf & 24.65 & 21.62 & 16.67 & 17.98 & 34.43 \\
        \hline
    \end{tabular}    
\end{table*}

The latency of the scan ($\sim$145 ms) and pose ($\sim$280 ms) topics is relatively high, but this is not necessarily due to the framework. The Lidar scan provides a header timestamp from the start of the full rotation, which takes 100 ms. The pose topic originates from SLAM, which depends on scan and odometry/global topics, and on the algorithm's processing requirements, which depend on the node's available resources.

\begin{figure}
    \centering
    \includegraphics[scale=0.54]{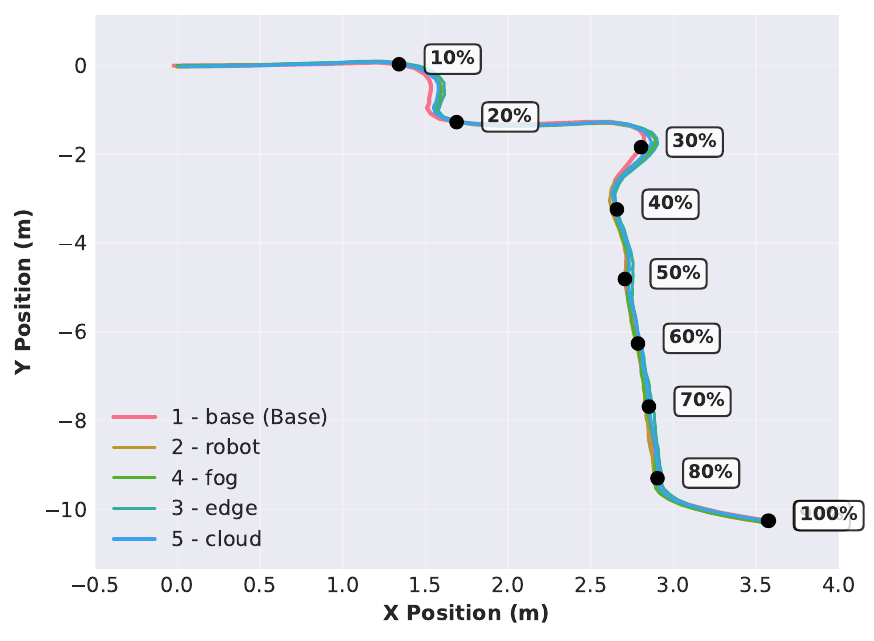}
    \caption{Spatial Map.\label{fig:map}}
\end{figure}

The resulting navigation for scenario 1 is presented in Fig. \ref{fig:map}, and a video is also available to show the robot behavior for each execution\footnote{ \url{https://youtu.be/4Bp8EqYhM5c}}. The base run is the one executed directly on the robot node without the BlazeAIoT framework. It is possible to see that the runs executed within the BlazeAIoT framework follow the shape of the base path, with a slight deviation. The error is more pronounced during rotations, which can be explained by network latency that slows the SLAM pose broadcast.

\begin{figure}
    \centering
    \includegraphics[scale=0.53]{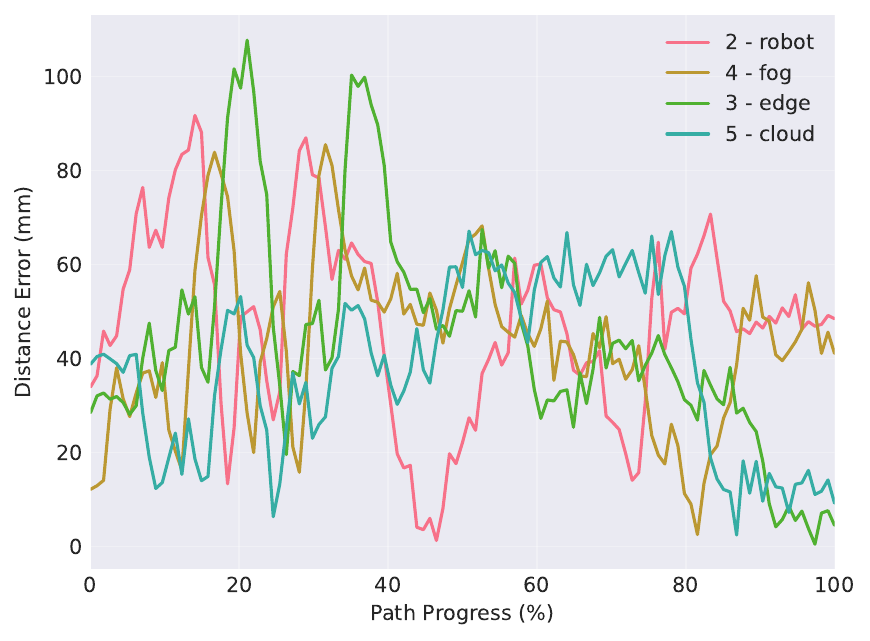}
    \caption{Total Distance Error (relative to Base).\label{fig:dist_err}}
\end{figure}

Fig. \ref{fig:dist_err} shows the Euclidean distance error relative to the base run. The edge run shows a greater error (up to 10.7 cm), which is the opposite of what would be expected. Still, the reason for that is the increased traffic and CPU load introduced by the addition of two subscribers (one on the robot node and one on the edge node) to be able to do the latency monitoring. The fog run shows lower error than the edge run (at most 8.5 cm), attributable to the fog node's low latency combined with high computational power. The robot run shows better performance than the fog run with an error of at most 7.3 cm. The cloud presents the highest error (up to 12.9 cm), but follows the same path as the other runs. The main reason for this increased error is the greater network latency (+30 ms). When the robot receives a command, it executes almost two control steps (at 50 Hz). The robot is always trying to catch up for the delay.

Unsurprisingly, the best allocation for this scenario is to run everything on the robot's edge node. The communication latency is negligible, and because the entire communication runs directly on ROS2 without a bridge, there is no overhead. This is true when the robot's node resources are sufficient. Adding more services to the robot may yield a different conclusion. This is the objective of the following scenario, in which the robot's node resources are pushed to their limits.

\subsection{Scenario 2: Emergency stop}
The second scenario is similar to that in \cite{Muratore2023XBot2D:Robotics} and a video is available to visualize the scenario execution\footnote{\url{https://youtu.be/PYek5ER0_MM}}. It involves moving the robot in a straight line and using hand gesture detection to detect a fully open hand, thereby triggering an emergency stop. The same four steps described for scenario 1 are executed. The goal is to evaluate the effect of latency in a high-volume, high-throughput scenario. The RGB images transmitted by the camera and shared through the BlazeAIoT framework are significantly influenced by the size of the image message packet. Additionally, the emergency stop service uses the AI model shown in Fig. \ref{fig:emergency_stop}, which is also resource-intensive relative to the other services. The relevant metric in this scenario is the time difference between the hand gesture's appearance and the robot's response.

To compare the network's effect on the robot's response, the emergency-stop ROS node was launched simultaneously across all layers. A monitoring service subscribes to the emergency stop topic, receives messages from all layers, and can compare their latencies. The layer nodes running the emergency stop ROS node are robot-1, edge-gpu2, fog-gpu3, and cloud-gpu2. It is important to note that the MediaPipe hand-gesture algorithm runs on the CPU, as a GPU-based version was not natively available on the Jetson. Still, the results highlight an interesting aspect for resource-constrained nodes. 

Table \ref{tab:estop_metrics} presents the metrics obtained during execution.

\begin{table*}
    \centering
    \caption{Emergency stop metrics}
    \label{tab:estop_metrics}
    \begin{tabular}{|c|c|c|c|c|c|c|}
        \hline
        Node & Image     & Image        & Processing & Emergency Stop & Difference  \\
             & Rate (Hz) & Latency (ms) & Time (ms)  & latency (ms)   & (\%) \\
        \hline
        robot-1 & 27 & 38.6 & 430 & 1287.3 & - \\
        \hline
        edge-gpu2 & 21 & 40.9 & 190 & 581.6 & -548 \\
        \hline
        fog-gpu3 & 7.9 & 44.7 & 23 & 181.3 & -859 \\
        \hline
        cloud-gpu2 & 7.9 & 172.5 & 19 & 550.74 & -572 \\
        \hline
    \end{tabular}
\end{table*}

The variation in image frame rate is due to rate-limiting logic that respects the allowed bandwidth budget. The camera images are the first topic to be bridged through BlazeFlow. The remaining messages are then added, including the scan, tf, encoder\_odometry, and imu topics, which together constitute the largest payloads after the camera images.

The service running on the robot receives images at 22 Hz, which is lower than the expected 30 Hz. The first justification for this difference is the node memory load. Since the Jetson shares memory with the CPU and GPU, the combination of the emergency stop process, BlazeFlow execution, and all robot processes places a significant memory demand, ranging from 97\% to 100\%. The result is that emergency stop service takes an average of 725 ms to execute on one image. Because the image callback is reentrant, multiple frames are processed in parallel (approximately 16 concurrent processes), thereby increasing the typical load. The other justification for the rate difference is based on the metrics-monitoring methodology, as in scenario 1. To be able to compute the topic latency, an additional subscriber is added on the robot node and on the edge node, which increases the network traffic and the necessary CPU on the robot since the multicast is not leveraged with the default ROS configuration, meaning that the robot node sends the topic message four times instead of 2, adding to the CPU burden.

The service running on the edge node receives the images at the same rate as the robot. Still, because the node is not running the robot and sensor services, the edge node's memory usage is approximately 85\%, resulting in a 190 ms process time. The emergency stop service remains slower than the input rate, resulting in an accumulation of process threads (approximately four concurrent processes).

To assess whether the hand gesture is detected, three consecutive positive detections are required. If the service is performed locally on the robot, without the network latency, it takes approximately 861 ms to generate the emergency stop signal. On the edge node, the emergency stop is triggered after 333 ms, whereas on the fog and cloud layers, it is triggered after 380 ms.

From the results, the emergency stop executes faster on the fog layer, with an 859\% reduction in time compared to running it on the robot directly, even though the image rate to the fog is about 26\% of the expected 30 Hz. The cloud, even with a reduced rate and higher latency, offers a 572\% reduction in time. Finally, remote execution of the emergency stop on a different edge node results in a 548\% decrease in execution time. While fog and cloud execution offer better response times, the frame-rate reduction could be problematic due to the risk of losing information from filtered frames.

It should be noted that the emergency stop service runs in all scenarios, creating a worst-case timing issue due to high CPU/memory usage. If the service is allocated to a remote node, overall performance would improve.

\section{Discussion}\label{section:discussion}
Across the two scenarios, it is evident that when a solution handles small messages, latency has little impact on the robot's behavior for non-critical tasks. If a fast reaction is required, the time added by the network latency can be problematic. For a solution that handles large messages and computationally intensive tasks, the ability to deploy the service to another layer can be beneficial. If the local node is running multiple services, limited CPU availability can increase processing time. While latency is higher in the other layers, a high-end CPU/GPU can be much faster than a typical edge node and compensate for the communication speed.

The BlazeAIoT framework facilitates communication by compressing payloads and rate-limiting, reducing the message rate to fit within the allowed bandwidth. Service allocation is essential, but orchestration is not part of the actual implementation. With BlazeFlow's dynamic capabilities, moving a service from one node to another is seamless, as all services broadcast their advertise/request topics.

\subsection{BlazeAIoT vs. Objectives}
After describing the BlazeAIoT components, it is possible to assess how the platform addresses the objectives enumerated in Section \ref{section:motivation}.

\begin{itemize}
    \item \textbf{Cost-aware solution}: all the technologies used in the BlazeAIoT platform are Open-Source technologies (MongoDB, Kafka, RKE2, K3S, DDS, Redis, and Prometheus). Using Rancher as a Kubernetes management tool facilitates the creation of lightweight Kubernetes clusters with K3S and full-scale Kubernetes clusters with RKE2. This solution can be deployed on any cloud provider or replaced with other Kubernetes services, such as Azure Kubernetes Service, Amazon Elastic Kubernetes Service, and Google Kubernetes Engine. The result is a cloud-agnostic solution that can easily deploy within an enterprise's existing infrastructure.
    \item \textbf{Programming language flexibility}: BlazeFlow data technologies were selected based on this objective. The platform's clients do not directly access DDS. They would rather connect to Kafka or Redis, which provide client libraries in multiple languages. ROS2 packages can still communicate directly in the same layer. A converter is natively included in BlazeFlow to translate between any broker and ROS2. Client services typically use the SDK, which is available in multiple languages, and it is straightforward to add support for new languages as needed.
    \item \textbf{Multi-layer support}: BlazeFlow enables communication in all layers. In Kubernetes, a virtual private network is available to all nodes within the cluster. To enable multi-layer, multi-cluster communication, IP tunneling, such as Submariner, bridge that gap through an encrypted tunnel.
    \item \textbf{Flexible service distribution}: With BlazeConfig, all services can fetch the layer, node, and service configuration. It is possible to have multiple instances of a service, each with its own configuration if needed.
    \item \textbf{Real-time performance}: BlazeFlow broker's performances were studied in a previous work \cite{Melancon2023BlazeFlow:Applications}. To monitor the metrics of the service within the platform, BlazeMonitor enables them for eventual improvement to place the services to ensure 
    the real-time execution.
    \item \textbf{Robustness and security}: The connectivity loss scenario was not covered by this work, but all the actual development makes it easy to spawn the required services locally if the resources are sufficient to take over while waiting for the connectivity to come back. Security is a topic in its own right, but all selected technologies offer secure options. For all the communications, it is possible to encrypt the data. MongoDB also offers data-at-rest encryption. Using Kubernetes limits access to its internal network and provides authorization and role-based access controls to help secure the entire solution. Traffic between layers is also encrypted using an IPSec tunnel. It would also be possible to add monitoring services within BlazeMonitor to collect metrics for this objective and to take action. But this was not part of the current implementation.
\end{itemize}

\section{Conclusion}\label{section:conclusion}
This work presented BlazeAIoT, a modular multi‑layer platform designed to unify distributed robotics across edge, fog, and cloud infrastructures. By extending the BlazeFlow framework, the proposed solution addresses critical challenges in distributed robotics, including dynamic data transfer, adaptive communication, real‑time performance, and resilience against connectivity loss. The integration of BlazeConfig for multi‑layer configuration management, BlazeFlow for dynamic and adaptive data distribution, and BlazeMonitoring for system health metrics provides a comprehensive foundation for robust and scalable deployments.

Experimental validation through navigation and AI‑driven message handling scenarios demonstrated the platform’s ability to meet stringent real‑time constraints while maintaining flexibility across distributed topologies. The results confirm that BlazeAIoT can dynamically allocate services, optimize bandwidth usage, and ensure interoperability across heterogeneous environments.

Beyond robotics, the platform’s design principles make it equally applicable to broader IoT domains, including smart cities, smart factories, and cyber‑physical systems. Future work will focus on enhancing load balancing in BlazeFlow, exploring more efficient inter‑cluster networking solutions, and proposing an orchestration solution to support service allocation. With these improvements, BlazeAIoT has the potential to serve as a foundation for distributed robotics and other distributed systems, such as IoT.

\bibliographystyle{unsrtnat}
\bibliography{references}  






\end{document}